\documentclass[10pt,twocolumn,letterpaper]{article}

\usepackage[pagenumbers]{iccv} 

\definecolor{iccvblue}{rgb}{0.21,0.49,0.74}
\usepackage[pagebackref,breaklinks,colorlinks,allcolors=iccvblue]{hyperref}
\usepackage[cal=boondox]{mathalfa}
\usepackage{multirow}
\usepackage{graphicx}
\usepackage{colortbl}
\usepackage{xcolor} 
\usepackage{subcaption}

\usepackage{tabularx}   
\usepackage{makecell}   
\usepackage{booktabs}   

\usepackage{booktabs}  
\usepackage{amssymb}   
\usepackage{pifont}    
\usepackage{xcolor}    
\usepackage{makecell}  

\newcommand\teaserfigure[1]{\gdef\@teaser{#1}}

\definecolor{baselinecolor}{gray}{.9}
\def\paperID{4148} 
\def\confName{ICCV}
\def\confYear{2025}

\title{$A_{0}$: An Affordance-Aware Hierarchical Model for General Robotic Manipulation}
\author{
Rongtao Xu\textsuperscript{1,4}\footnotemark[1]\:\;\footnotemark[3],
Jian Zhang\textsuperscript{1}\footnotemark[1],
Minghao Guo\textsuperscript{1}\footnotemark[1],
Youpeng Wen\textsuperscript{2}\footnotemark[1],
\\
Haoting Yang\textsuperscript{3},
Min Lin\textsuperscript{2},
Jianzheng Huang\textsuperscript{3},
Zhe Li\textsuperscript{3}, Kaidong Zhang\textsuperscript{2}, Liqiong Wang\textsuperscript{3}, \\ Yuxuan Kuang\textsuperscript{5}, Meng Cao\textsuperscript{1}, 
Feng Zheng\textsuperscript{3}\footnotemark[2],
Xiaodan Liang\textsuperscript{1,2}\footnotemark[2]
\\
\textsuperscript{1·}MBZUAI  
\textsuperscript{2}SYSU 
\textsuperscript{3}SUSTech 
\textsuperscript{4}Spatialtemporal AI 
\textsuperscript{5}CMU 
\\
\color{red}{https://a-embodied.github.io/A0/}
}

\begin{document}
\maketitle
\renewcommand{\thefootnote}{\fnsymbol{footnote}} 
\footnotetext[1]{Equal contribution}
\footnotetext[2]{Corresponding authors} 
\footnotetext[3]{Project Leader}

\begin{abstract}

Robotic manipulation faces critical challenges in understanding spatial affordances—the ``where'' and ``how'' of object interactions—essential for complex manipulation tasks like wiping a board or stacking objects. Existing methods, including modular-based and end-to-end approaches, often lack robust spatial reasoning capabilities. Unlike recent point-based and flow-based affordance methods that focus on dense spatial representations or trajectory modeling, we propose $A_0$, a hierarchical affordance-aware diffusion model that decomposes manipulation task into high-level spatial affordance understanding and low-level action execution. $A_0$ leverages the Embodiment-Agnostic Affordance Representation, which captures object-centric spatial affordances by predicting contact point and post-contact trajectories. $A_0$ is pre-trained on 1 million contact points data and fine-tuned on annotated trajectories, enabling generalization across platforms. Key components include Position Offset Attention for motion-aware feature extraction and a Spatial Information Aggregation Layer for precise coordinate mapping. The model’s output is executed by the action execution module. Experiments on multiple robotic systems (Franka, Kinova, Realman and Dobot) demonstrate $A_0$'s superior performance in complex tasks, showcasing its efficiency, flexibility, and real-world applicability.
\end{abstract}
    
\section{Introduction}
\label{sec:intro}
\begin{figure}[!ht]
    \centering      \includegraphics[width=1\linewidth]{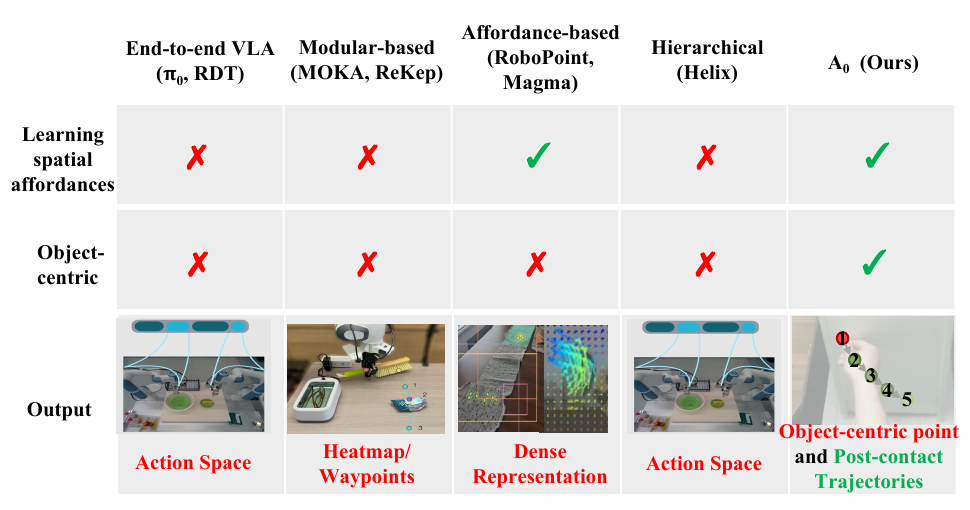}
    \caption{Comparison of different manipulation methods. $A_0$ is an object-centric hierarchical model that learns Embodiment-Agnostic Affordance Representation.}
    \label{fig:teaser}
\vspace{-0.6cm}
\end{figure}
Robotic manipulation is a fundamental yet challenging task in robotics and embodied AI, requiring robots to interact with objects in complex environments. Recent advancements focus on two approaches: (1) modular-based methods~\cite{kuang2024ram,huang2024rekep,liu2024moka} that use large vision foundation models for spatial understanding, and (2) end-to-end Vision-Language-Action (VLA) methods~\cite{brohan2023rt2,kimopenvla,black2024pi_0,liu2024rdt} for fine-grained manipulation. However, existing methods still face significant limitations in understanding spatial affordances—the ``where'' and ``how'' of object interactions—which are critical for achieving spatial intelligence. For example, in tasks like wiping a whiteboard, insufficient understanding of spatial affordances often leads to incomplete or inefficient execution.

\begin{figure*}[ht]
    \centering   
    \includegraphics[width=0.95\linewidth]{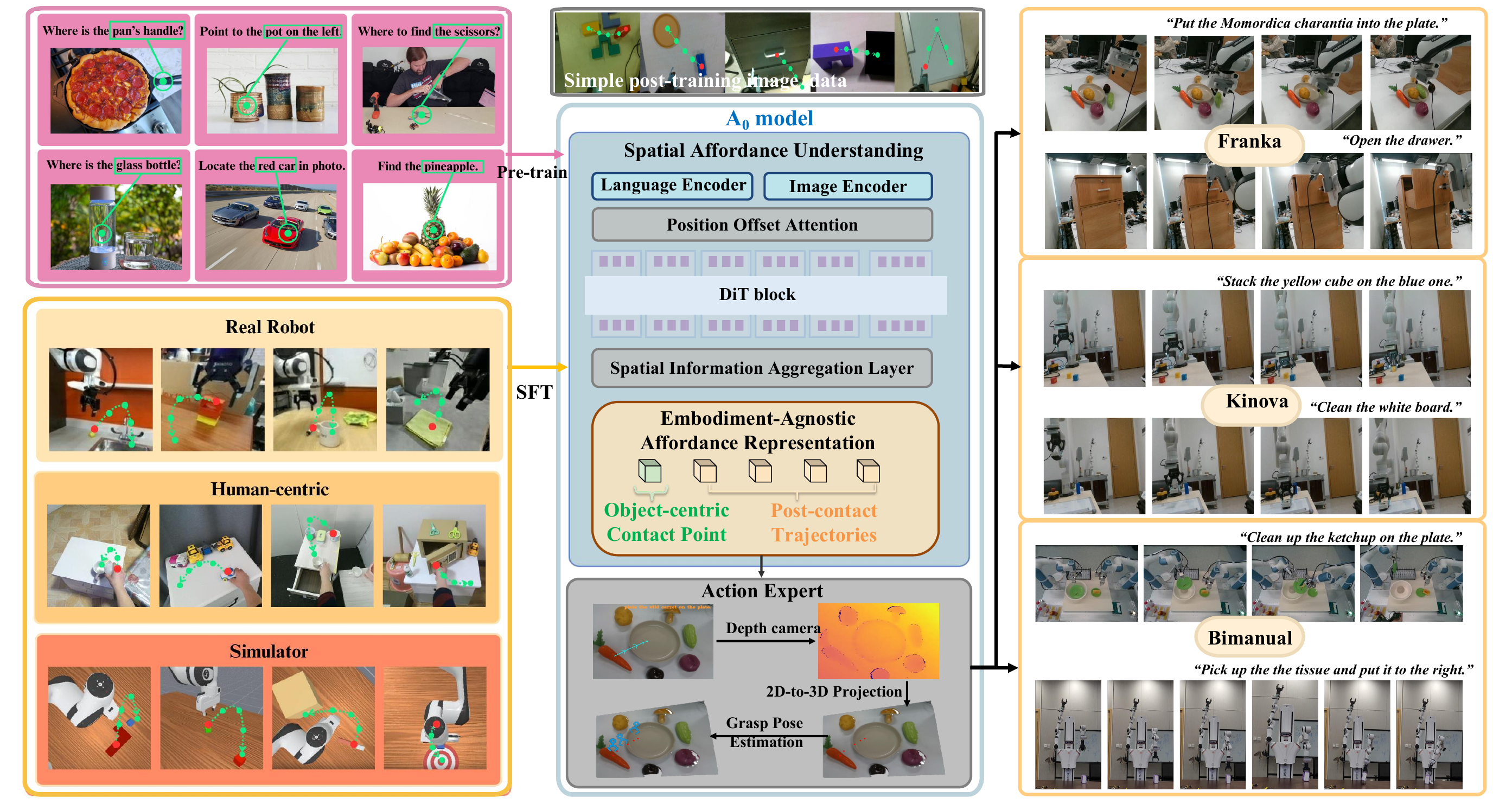}
    \caption{The \( A_0 \) model decomposes robotic manipulation tasks into two levels: (1) high-level spatial affordance understanding and (2) low-level action execution. \( A_0 \) leverages an Embodiment-Agnostic Affordance Representation to predict object-centric contact points and post-contact trajectories. The architecture includes well-designed key components for affordance learning. \( A_0 \) is pre-trained on a large-scale dataset of contact points and fine-tuned on annotated trajectories, enabling generalization across diverse robotic platforms. Zoom-in for the best of views. }
    \label{fig:1}
\vspace{-0.4cm}
\end{figure*}

Spatial affordances can be learned not only from real-world and synthetic robotic datasets~\cite{tao2024maniskill3} but also from a wide range of out-of-domain data rich in actionable knowledge, such as Internet data~\cite{deitke2024molmo}, and hand-object interaction (HOI) data~\cite{yuan2024general}. These datasets contain valuable information about object interactions, spatial properties, and physical attributes, making it essential to represent actionable knowledge as unified spatial affordances. Modular-based methods, such as ReKep \cite{huang2024rekep} and MOKA \cite{liu2024moka}, directly utilize Large Vision Models (LVM) but lack a deep understanding of the spatial and physical world, particularly in capturing the operability of objects. On the other hand, end-to-end methods like RDT \cite{liu2024rdt} and $\pi_0$ \cite{black2024pi_0} generate actions directly without adequately understanding spatial positions, leading to suboptimal performance in complex manipulation tasks, such as wiping a board or stacking objects.

Recently, several methods have emerged, recognizing the importance of spatial affordance in robotic manipulation. Point-based methods such as SpatialVLA~\cite{qu2025spatialvla}, RoboPoint \cite{yuanrobopoint}, Track2Act~\cite{bharadhwaj2024track2act}, and flow-based methods such as General Flow~\cite{yuan2024general} and Im2Flow2Act~\cite{xuflow} have made significant strides in modeling spatial interactions. However, these methods often focus on dense spatial representations or trajectory modeling (see Figure~\ref{fig:teaser}), which can be computationally expensive and embodiment-specific. In contrast, our approach is \textbf{object-centric}, focusing solely on predicting the contact point and trajectories of the objects that need to be manipulated. We propose an \textbf{Embodiment-Agnostic Affordance Representation} to capture the ``where'' and ``how'' of object interactions. This design makes our method \textbf{embodiment-agnostic}, enabling seamless generalization across different robotic platforms. By leveraging this general Embodiment-Agnostic Affordance Representation, our approach requires only a small amount of task-specific annotated data for fine-tuning, making it highly practical and general for real-world deployment.

To address the challenge of spatial understanding and physical reasoning in manipulation, we propose $A_0$, a novel \textbf{A}ffordance-Aware Hierarchical Model specifically designed for robotic manipulation. The model decomposes manipulation tasks into two levels: (1) a high-level \textbf{spatial affordance understanding} and (2) a low-level \textbf{action execution}. Our model and system architecture are depicted in Figure~\ref{fig:1}. $A_0$ primarily focuses on high-level spatial affordance understanding, including object contact points and post-contact trajectories, to guide low-level action execution effectively.
To learn foundational localization capabilities, $A_0$
  is pre-trained on 1 million contact-point localization data points, followed by supervised fine-tuning on an annotated spatial trajectory dataset. This hierarchical design enables $A_0$  to handle complex manipulation tasks more effectively, particularly those requiring spatial affordance reasoning and physical interaction. Our method achieves an average success rate of 62.50\% on Franka and 53.75\% on Kinova, outperforming the strongest baselines on both platforms. Notably, it demonstrates robust performance in trajectory-following tasks like Wipe Board (45\%).

The key contributions of our work are as follows:

\begin{itemize}
\item We introduce an \textbf{Embodiment-Agnostic Affordance Representation} that efficiently captures spatial affordance by predicting object-centric contact point and trajectories. This representation is supported by a 1 million annotated dataset and an efficient annotation pipeline. The point-based nature of our representation makes it highly versatile and easy to deploy across different robotic platforms.

\item We propose a \textbf{hierarchical affordance-aware diffusion model $A_0$} that first learns Embodiment-Agnostic Affordance Representation and then generates precise manipulation actions. The model incorporates several key components, including Position Offset Attention, DiT blocks, and a Spatial Information Aggregation Layer, to enhance spatial affordance understanding.

\item We validate the effectiveness of $A_0$ on \textbf{multiple robotic platforms} (e.g., Franka, Kinova, Realman and Dobot). The model demonstrates superior performance in complex tasks requiring spatial affordance reasoning, such as wiping a whiteboard or placing objects, showcasing its strong generalization capabilities and embodiment-agnostic design.

\end{itemize}

\begin{figure*}[t]
  \centering \includegraphics[width=0.9\textwidth]{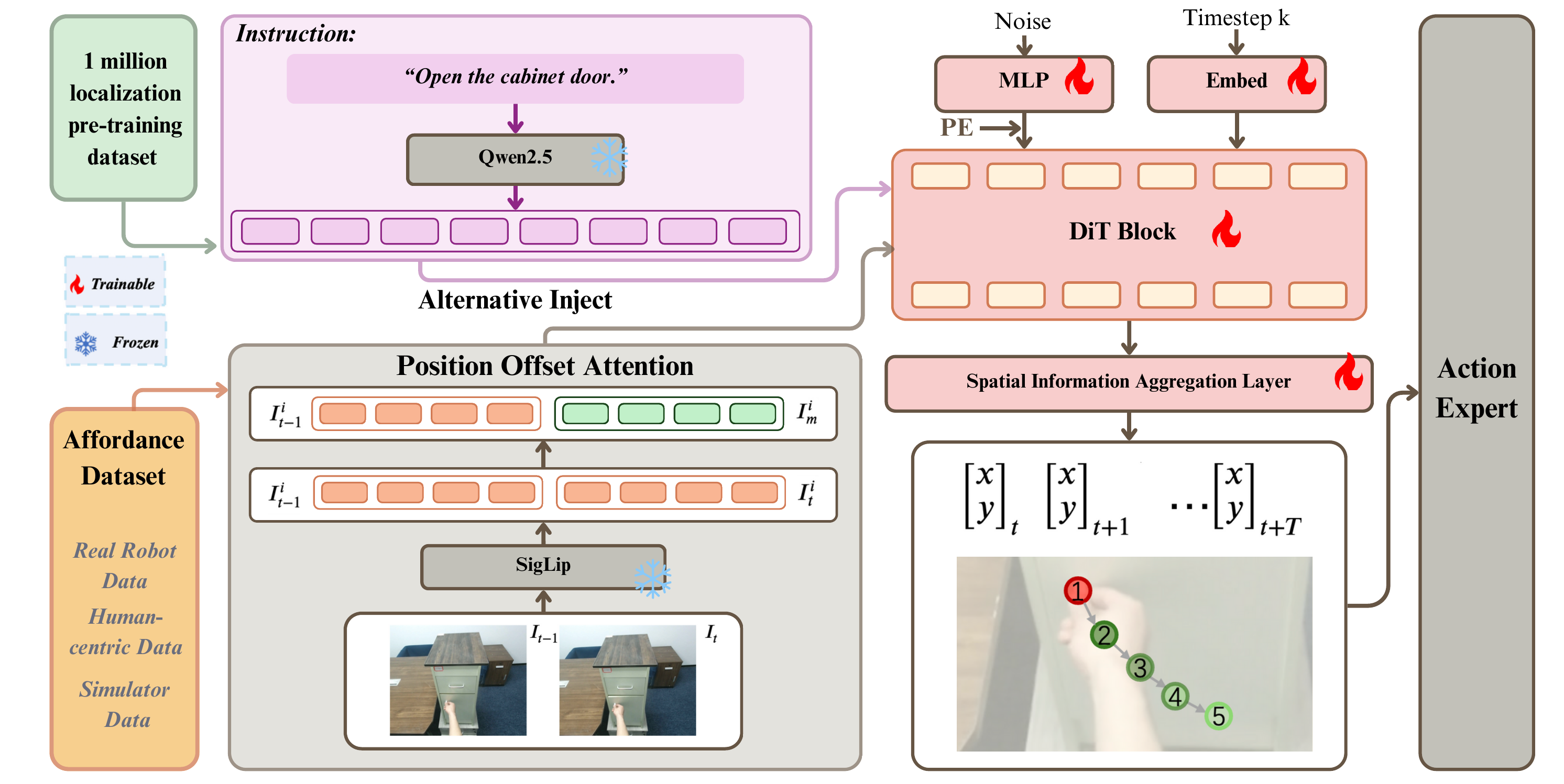} 
  \caption{Overview of $A_0$ model. The model is transformer based diffusion probabilistic model to predict the waypoints for robotic manipulation. We use the pre-trained Qwen2.5-7B~\cite{yang2024qwen2} and SigLip (400M)~\cite{zhai2023sigmoid} to encode the language instruction and images, separately. The image of previous time step are used to provide motion information by the proposed motion token enhancement. The image and text tokens are alternatively injected as conditions via cross attention.} 
  \label{fig:2} 
  \vspace{-0.4cm}
\end{figure*}

\section{Related Works}

\subsection{Spatial Affordance in Robotic Manipulation.}

Affordance prediction is crucial for robotic manipulation, enabling structured action selection in complex environments. Current methods vary in input processing: point cloud-based approaches \cite{moWhere2ActPixelsActions2021a, wuVATMartLearningVisual2022, wangAdaAffordLearningAdapt2023, gengEndtoEndAffordanceLearning2022, liLASOLanguageGuidedAffordance2024, lingArticulatedObjectManipulation2024} excel in spatial reasoning, while RGB image-based techniques \cite{bahlAffordancesHumanVideos2023, qianAffordanceLLMGroundingAffordance2024, chenAffordanceGroundingDemonstration2023, huangManipVQAInjectingRobotic2024, liuMOKAOpenWorldRobotic2024, huangA3VLMActionableArticulationAware2024, huangReKepSpatioTemporalReasoning} prioritize accessibility. Language integration, as in LASO \cite{liLASOLanguageGuidedAffordance2024} and Chen \textit{et al.} \cite{chenAffordanceGroundingDemonstration2023}, further enhances these methods.

Output representations differ in three ways: 1) \emph{Heatmap-based} methods \cite{moWhere2ActPixelsActions2021a, wuVATMartLearningVisual2022, wangAdaAffordLearningAdapt2023, gengEndtoEndAffordanceLearning2022, bahlAffordancesHumanVideos2023, liLASOLanguageGuidedAffordance2024, qianAffordanceLLMGroundingAffordance2024, lingArticulatedObjectManipulation2024, chenAffordanceGroundingDemonstration2023} localize affordances but are computationally intensive; 2) \emph{Bounding box} approaches \cite{huangManipVQAInjectingRobotic2024, liuMOKAOpenWorldRobotic2024, huangA3VLMActionableArticulationAware2024} balance efficiency and accuracy; 3) \emph{Keypoint-based} techniques \cite{liuMOKAOpenWorldRobotic2024, huangReKepSpatioTemporalReasoning} provide robust representations for articulated objects. Recent advances like MOKA \cite{liuMOKAOpenWorldRobotic2024} and A3VLM \cite{huangA3VLMActionableArticulationAware2024} refine keypoint and bounding box methods. Our work improves keypoint-based reasoning by predicting contact points and trajectories, offering precise, efficient action specifications for robotic applications.

\subsection{VLA and Hierarchical Models} 
Recent advancements in Vision-Language-Action (VLA) models have enhanced robotic tasks through multimodal learning, categorized into transformer-based, VLM-based, and diffusion-based approaches~\cite{zhang2024navid,xu20253d,zhang2025robridge,ma2025phyblock,han2025multimodal}. Transformer-based methods \cite{wuUnleashingLargeScaleVideo2023, cheangGR2GenerativeVideoLanguageAction2024, bharadhwajGen2ActHumanVideo2024} predict action sequences using video generation models. VLM-based approaches \cite{quSpatialVLAExploringSpatial2025, liCogACTFoundationalVisionLanguageAction2024, kimopenvla, brohanRT2VisionLanguageActionModels2023, zhen3DVLA3DVisionLanguageAction2024, chenPaLIXScalingMultilingual2023, awadallaOpenFlamingoOpenSourceFramework2023} leverage pre-trained vision-language models for improved generalization. Diffusion-based frameworks \cite{houDiffusionTransformerPolicy2025, liuRDT1BDiffusionFoundation2024, teamOctoOpenSourceGeneralist2024, reuss2023goalconditioned, pearce2023imitating, wen2024vidmanexploitingimplicitdynamics} enable probabilistic action generation for robust trajectory planning. GR-2 \cite{cheangGR2GenerativeVideoLanguageAction2024} and OpenVLA \cite{kimopenvla} exemplify transformer-based and VLM-based methods, respectively, while RDT-1B \cite{liuRDT1BDiffusionFoundation2024} showcases diffusion-based bimanual manipulation. These models leverage large datasets to enhance real-world robotic applications.

Hierarchical models enhance manipulation capabilities by organizing control into multiple abstraction levels, enabling robots to generalize across diverse tasks. DexGraspNet~\cite{wang2022dexgraspnet} provides a large-scale dataset and learning framework for hierarchical dexterous grasping, combining high-level planning with low-level control to achieve robust grasps across various object categories. Similarly, Helix~\cite{figure_ai_helix} adopts a hierarchical reinforcement learning paradigm, leveraging high-level policies to guide low-level motor control, improving both adaptability and efficiency in real-world manipulation scenarios. 


Unlike VLA methods, which often struggle with embodiment-specific constraints, $A_0$ leverages a diffusion-based framework for flexible and robust trajectory planning under uncertainty. Compared to hierarchical models, $A_0$ adopts a hierarchical affordance-aware design that decomposes tasks into high-level spatial affordance understanding and low-level action execution, enabling generalization across diverse tasks and platforms. In contrast, Helix leverages vision-language models (VLMs) to learn latent semantic representations rather than explicit object-centric spatial affordances. By focusing on object-centric spatial affordances, $A_0$ reduces computational overhead and improves adaptability, making it highly practical for real-world deployment.


\section{The $A_{0}$ Model}

\subsection{Overview}
 
We propose the $A_0$ model to address spatial affordance understanding (the ``where'' and ``how'' of object manipulation). As shown in Figure~\ref{fig:2}, $A_0$ uses an Embodiment-Agnostic Affordance Representation (See Sec.~\ref{sec:eaar}) to predict object contact points and trajectories, supported by a multi-dataset annotation pipeline. The model operates hierarchically: (1) high-level spatial affordance understanding and (2) low-level action execution. For spatial understanding, we leverage Diffusion architecture for object affordance learning (See Sec.~\ref{sec:Structure}). Furthermore, the training strategy includes pre-training on 100,000 contact-point samples and fine-tuning on annotated trajectories (see Sec.~\ref{sec:train}). For action execution, we adopt point-based methods similar to~\cite{liu2024moka,kuang2024ram} (see Sec.~\ref{sec:action}).

\subsection{Embodiment-Agnostic Affordance Representation}
\label{sec:eaar}

To enable the model to effectively understand spatial affordances and capture the ``where'' and ``how'' of object interactions, we propose a unified \textbf{Embodiment-Agnostic Affordance Representation} \(\mathcal{R}\) that can be easily acquired from diverse data sources. Specifically, this representation integrates actionable knowledge from real-world or synthetic robotic data \(\mathcal{R}_{\text{R}}\), hand-object interaction (HOI) data \(\mathcal{R}_{\text{H}}\), and custom data \(\mathcal{R}_{\text{C}}\). To unify affordance information from these varied sources, each entry in the Embodiment-Agnostic Affordance Representation includes an object-centric RGB image \(I\) (the current frame), a set of 2D waypoints indicating the contact point \(c_0^{2\text{D}}\) and post-contact trajectories \(T = (t_0^{2\text{D}}, t_1^{2\text{D}}, t_2^{2\text{D}},\cdots)\), along with a manipulation task instruction expressed in natural language. Therefore, the constructed Embodiment-Agnostic Affordance Representation can be expressed as:
\begin{flalign}
\begin{aligned}
&\mathcal{R} = \mathcal{R}_{\text{R}} \cup \mathcal{R}_{\text{H}} \cup \mathcal{R}_{\text{C}}  \\
&= \{(I,L,C,T) \mid C = (c_0^{2\text{D}}), T = (t_0^{2\text{D}}, t_1^{2\text{D}}, t_2^{2\text{D}},\cdots)\}.
\end{aligned}
\end{flalign}

By leveraging this general Embodiment-Agnostic Affordance Representation, our model requires only a small amount of task-specific annotated images for fine-tuning, enabling rapid cross-platform deployment. This design ensures that the model remains object-centric, focusing solely on predicting the contact point position and trajectories of the objects that need to be manipulated, while maintaining computational efficiency and embodiment-agnostic flexibility. For different data sources, we employ distinct strategies to extract and annotate the Embodiment-Agnostic Affordance Representation.

\textbf{Dataset:} The dataset comprises four components: PixMo-One-Point, HOI4D-22k, DROID-3k, and Maniskill-5k. PixMo-One-Point includes one million single-contact-point annotations from PixMo-Points~\cite{deitke2024molmo}. 
HOI4D-22k contains 22,000 human-object interaction trajectories extracted and verified from HOI4D~\cite{liu2022hoi4d}.
DROID-3k consists of 3056 verified manipulation trajectories from the DROID dataset~\cite{khazatsky2024droid}, annotated using a semi-automated pipeline.  
Maniskill-5k includes 4965 trajectories from the ManiSkill Scene~\cite{taomaniskill3} dataset, converted to 2D for compatibility. For detailed descriptions of the dataset and annotation pipeline, see Appendix.

\subsection{$A_{0}$ Model Structure}
\label{sec:Structure}
Figure \ref{fig:2} provides an overview of the proposed $A_0$ model. Our approach builds upon the diffusion transformer architecture, DiT \cite{peebles2023scalable}. In our framework, a pre-trained vision encoder and a pre-trained text encoder are employed to extract features from the input image and language instruction, respectively. The resulting image and text tokens are subsequently integrated via a cross-attention mechanism, which conditions the diffusion process.

We define $T$ waypoints as affordance representation
\begin{equation}
\mathbf{x}_{t:t+T}, \quad \text{where} \quad \mathbf{x}_t = (u, v) \in [0,1]^2 \subset \mathbb{R}^2.
\end{equation}
$t$ denotes the $t$-th timestep in the trajectory. $T$ is the future step chunk sizes. $ \mathbf{x}_t $ denotes a two-dimensional coordinate normalized according to the image size.
The first point is start point, indicating the initial contact position.

The input to the $A_0$ model comprises diffusion timestep $k$ and noisy waypoints.
Followed by DiT, the vector embedding of $k$ is appended as additional tokens in the input sequence. The input observation images $I_{t-1:t}$ and language instruction $\ell$ are incorporated as conditions via cross-attention. The observed images captured from an external camera consist of the current frame $I_t$ and the previous frame $I_{t-1}$. Optionally, the previous frame can be utilized to provide motion information between the images. 

We use the vision encoder of pre-trained SigLiP (400M) \cite{zhai2023sigmoid} to separately encode the observation images $I_{t-1},I_t$ to tokens  $I_{t-1}^1,I_{t-1}^2,...,I_{t-1}^{N_S}$ and $I_t^1,I_t^2,...,I_t^{N_S}$, where $N_T$ is the number of tokens.

\textbf{Position Offset Attention.} The motion information of objects is crucial for robotic manipulation. To enable the model to pay attention on motion information, we subtract the tokens \( I_t^i \) and \( I_{t-1}^i \) to obtain \( I_m^i \), which is then concatenated with \( I_t \) along the token dimension to serve as the final visual feature $o_t = concat([I_t^i, I_m^i], dim=1)$.

Sinusoidal positional embeddings in multi-dimensional grids \cite{liu2024rdt} are added to enhance the model’s ability to distinguish images based on viewpoint and time steps. We utilize pre-trained Qwen2.5-7B \cite{yang2024qwen2} as a robust language encoder to get tokens and masks. The image and text tokens are alternately injected in successive layers' cross-attention blocks to compress and address the different dimensions \cite{liu2024rdt}. The whole model consists of N layers of DiT blocks.  Our model learns a conditional distribution $p(\mathbf{x}_t|\ell, I_{t-1:t})$.

\textbf{Spatial Information Aggregation Layer.} We add a final nonlinear MLP decoder as a projection from the latent space
back to the physical space.

At inference stage, at timesteps $t$, we have language instruction $\ell$ and images $I_{t-1:t}$. we sample noisy waypoints $\mathbf{x}_{t:t+T}^k \sim \mathcal{N}(0, \mathbf{I})$. With diffusion timesteps $k$ and condition images and language instruction, the model can predict diffusion policies of waypoints $\mathbf{x}_{t:t+T}$ via fast ordinary differential equation (ODE) slover \cite{lu2022dpm} in $K_D$ de-noising steps ($K_D \ll K_F$). Denote $\mathbf{x}_{t:t+T}^k$ as $\mathbf{x}_k$, and thus the diffusion ODE is
\begin{equation}
    \frac{\mathrm{d}\mathbf{x}^k}{\mathrm{d}k}:=f(k)\mathbf{x}^k+\frac{g^2(k)}{2\sigma_k}\epsilon_\theta(\mathbf{x}^k,k), \mathbf{x}^k\sim\mathcal{N}(0,\tilde{\sigma}^2\boldsymbol{I}),
\end{equation}

where $f(k)\mathbf{x}_k$ is a linear function of $\mathbf{x}_k$, and $\frac{g^2(k)}{2\sigma_k}\epsilon_\theta(\mathbf{x}_k,k)$ is generally a nonlinear function of $x_t$ because of the neural network $\epsilon_\theta(\mathbf{x}^k,k).$
\par
\par





\subsection{Training Affordance Prediction Model $A_{0}$}
\label{sec:train}
\textbf{Pre-training.}
The prediction of the first waypoint (start point) is crucial; subsequent waypoints can be derived by learning the offset $\Delta x$. The model must be capable of localizing objects based on textual descriptions. Using these images of PixMo-One-Point sourced from the Internet, we trained the $A_0$ model to develop a general object localization capability.

To adapt to this dataset, we only use one image $I_t$ for the vision encoder and supervise the first waypoint $x_t^k$, which consists of normalized two-dimensional coordinates. The loss is MSE of ground truth and predicted coordinates:
\begin{equation}
    \mathcal{L}_p(\theta)=\frac{1}{n} \sum_{i=1}^{n} ((x_t^0)_i - (f_{\theta}(k,x_{t}^k,I_{t},\ell))_i)^2.
\end{equation}

\textbf{Supervised fine-tuning.} To adapt the pre-trained $A_0$ model to the robotic manipulation, we fine-tune the model on labeled data of specific task. In this stage, the text condition is extended from object label to language instruction. The output is extended from one point to $T$ waypoints. The model learns dynamic robotic manipulation with motion information.

Given the ground truth waypoints $x_{t:t+T}^0$, for timesteps $k \in K_F \subset \mathrm{N}^{+}  $ of diffusion forward process, we can add noise to the ground truth and sample $x_{t:t+T}^k=\sqrt{\bar{\alpha}^{k}} x_{t:t+T}^0+\sqrt{1-\bar{\alpha}^{k}} \epsilon^{k}$, where $\epsilon^{k} \sim \mathcal{N}(0, \mathbf{I})$, and $\bar{\alpha}^{k}$ is hyperparameter defined by the noise schedule. Given timesteps k, noisy waypoints $x_{t:t+T}^k$ and conditions $I_{t-1:t}$ and $\ell$, our model $f_{\theta}$ learns to prediction the origin waypoints and minimize the mean-squared error (MSE) loss: 
\begin{multline}
    \mathcal{L}_s(\theta) = \\ \frac{1}{n} \sum_{i=1}^{n} \bigl((x_{t:t+T}^0)_i 
- (f_{\theta}(k, x_{t:t+T}^k, I_{t-1:t}, \ell))_i \bigr)^2.
\end{multline}

\section{Action Execution}
\label{sec:action}
To execute the predicted actions from our model, we transform the predicted 2D keypoints $x_{t:t+T}$ into 3D space and determine the grasp pose using sampling based affordance lifting methods~\cite{fang2020graspnet, kpam}.

\subsection{2D-to-3D Projection}
Given the predicted 2D keypoints $x_{t:t+T}$ from model $A_0$, where $\mathbf{x}_t$ corresponds to the contact point and the remaining $T-1$ points represent post-contact directional cues, we employ depth-based deprojection to map them to 3D coordinates. The transformation is defined as:
\begin{equation}
    X_i = D(\mathbf{x}_i) K^{-1} \tilde{\mathbf{x}}_i, \quad i = t+1, t+2, \dots, t+T,
\end{equation}
where $D(\mathbf{x}_i)$ is the depth value at pixel $\mathbf{x}_i$, $K$ is the camera intrinsic matrix, and $\tilde{\mathbf{x}}_i$ represents the homogeneous coordinates of $\mathbf{x}_i$.

\subsection{Grasp Pose Estimation}
Following MOKA~\cite{liu2024moka} and RAM~\cite{kuang2024ram}, we refine the grasp pose by querying GraspNet~\cite{fang2020graspnet} or other grasp samplers~\cite{kpam}. They generates a set of grasp candidates based on local geometric features. We select the grasp candidate $G^*$ closest to the projected grasp point $X_t$:
\begin{equation}
    G^* = \arg\min_{G \in \mathcal{G}} \| G - X_t \|,
\end{equation}
where $\mathcal{G}$ is the set of grasp candidates proposed by grasp samplers.

\subsection{Waypoint Selection and Execution}
For waypoint execution, we leverage the post-contact directional keypoints $\{x_2, \dots, x_T\}$, which are projected into 3D space using the same depth deprojection method. To determine their heights in free space, we adopt a discrete selection strategy inspired by~\cite{yang2023setofmark}. The VLM is prompted to choose the height category (e.g., at target level or above target), and the final 3D waypoints are sampled accordingly.

The obtained grasp pose and 3D waypoints are then used to generate the motion trajectory in SE(3) space, ensuring smooth and feasible execution on the real robot.




\section{Experiment}
The primary goal of our experiments is to validate $A_0$'s architecture and analyze the behavior of $A_0$ on real-world tasks with a wide range of objects and tasks. We design our experiments to investigate the following questions:
\begin{itemize}
\item \textit{Is our model architecture effective for point-based affordance prediction? Does the pre-training improve performance on specific tasks?}
\item \textit{How does our approach perform in real-world scenarios? Is it platform-agnostic and transferable to other robotic systems?}
\item \textit{How does our model compare to state-of-the-art VLM-based policies and VLA policies?}
\end{itemize}

To address these questions, we design two types of experiments. First, we determine an appropriate network architecture and pretraining strategy using offline MAE evaluation metrics in Section \ref{sec:pretrain}. Then, we deploy our model on multiple robotic platforms and compare its performance with state-of-the-art methods in Section \ref{sec:realworld}.

\subsection{Offline Validation}
we employ the Mean Absolute Error (MAE) of ground truth and predicted waypoints to evaluate the performance of our model.
\begin{equation}
\text{MAE} = \frac{1}{n} \sum_{i=1}^{n} |(\mathbf{x}_{t:t+T})_i - (\mathbf{\hat{x}}_{t:t+T})_i|
\end{equation}

\label{sec:pretrain}
\subsubsection{Experimental Setting}
We configure the number of transformer layers N=28, and the model contains 1 billion parameters (denoted as $A_0$-1B). We set the chunk size T=5, and set the number of steps for the forward and backward processes of the diffusion model, $K_F$ and $K_D$, to 1000 and 5, respectively.
We split the DROID-3k, HOI4D-22k, and ManiSkill-5k datasets into training and testing sets with an 8:2 ratio. 3,600 images from the PixMo-One-Point dataset are used as the test set.
During the pre-training phase, we trained the model for 80,000 steps over 5 days. For the supervised fine-tuning, we train the model for 30000 steps,taking 50 hours. We utilized four A100 80GB GPUs with a batch size of 200. The 1B model required 73 GB of memory per card.

\subsubsection{Effectiveness of Pre-training}
The pre-training stage of $A_0$ is crucial for establishing foundational object localization capabilities. By training on 1 million contact-point localization samples, $A_0$ learns to predict initial contact points based on textual instruction. We employed Real-to-Sim and Sim-to-Real paradigms, which correspond to training on HOI4D and DROID with testing on Maniskill, and training on Maniskill with testing on HOI4D and DROID, respectively.

\begin{figure}[!ht]
    \centering
    \includegraphics[width=0.9\linewidth]{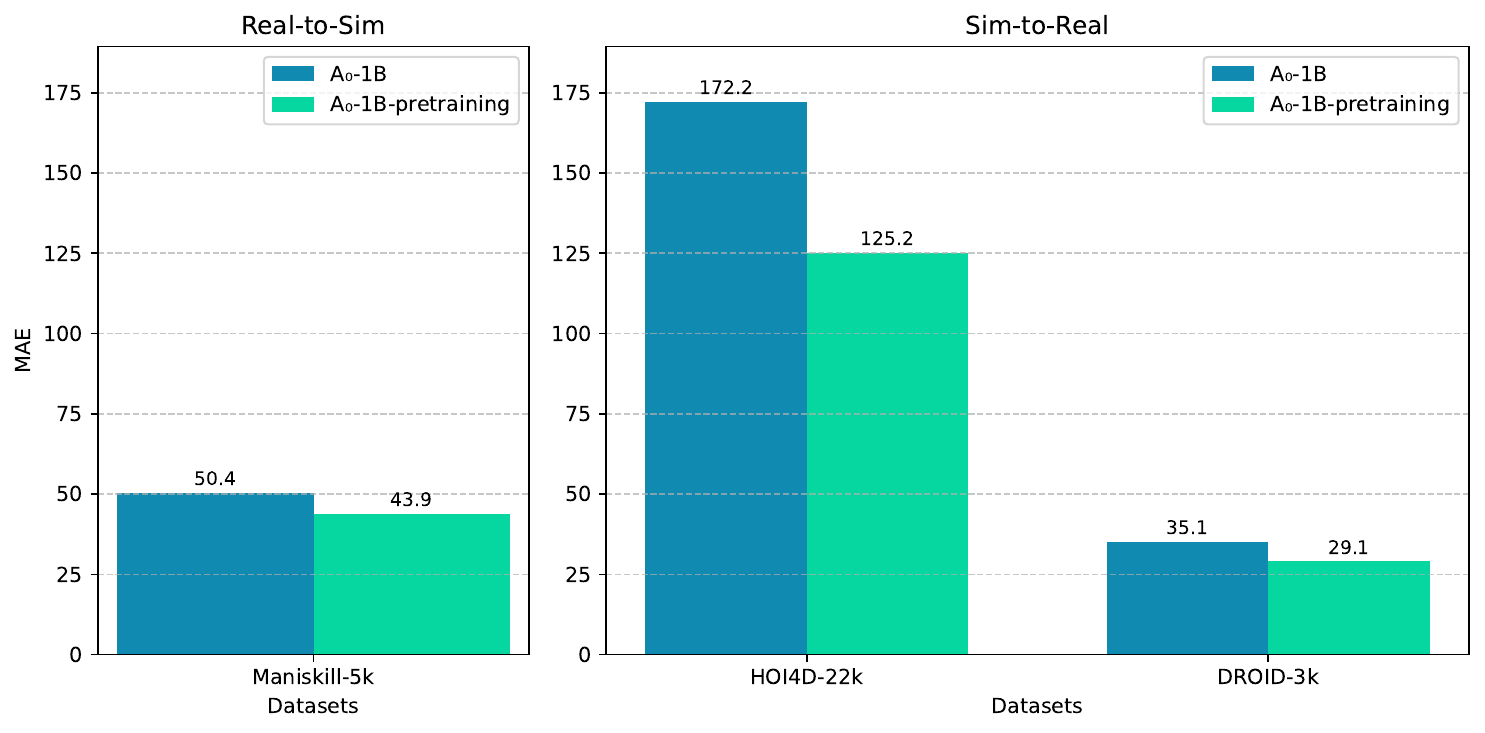}
    \caption{Performance of MAE$\downarrow$ with pretraining on three datasets.}
    \label{fig:pre}
    \vspace{-0.5cm}
\end{figure}

As shown in Figure \ref{fig:pre}, pre-training can decrease the MAE of $T$ waypoints on specific tasks. Ablation studies demonstrate that pre-training enhances $A_0$'s ability to generalize across unseen objects and environments. 
This highlights the importance of large-scale pre-training in building robust foundational capabilities for downstream robotic manipulation tasks.

\subsubsection{Effectiveness of Network Structure}
Table \ref{tab:network} show that Position Offset Attention and Spatial Information Aggregation Layer are crucial for A0’s performance. Removing Position Offset Attention increases the MAE by 0.8 on Maniskill-5k, impairing motion-aware reasoning, especially in tasks like wiping or stacking. Without the Spatial Information Aggregation Layer, the MAE increases by 13.2 pixels on HOI4D-22k, significantly degrading waypoint prediction in complex or occluded environments.

\begin{table}[!ht]
    \centering
    \caption{Ablation studies of network architecture. $A_0$-1B is pretrained on Pixmo-One-Point. 'POA' denotes Position Offset Attention and 'SIAL' denotes Spatial Information Aggregation Layer. We use MAE (lower is better) as evaluation metric.}
    \label{tab:network}
    \begin{tabular}{l@{}|c@{}|c@{}|c} 
        \toprule
        MAE$\downarrow$ & HOI4D-22k & Maniskill-5k & DROID-3k \\ \midrule
        $A_0$-1B & 47.5  & 5.5 & 17.5 \\ 
        $A_0$-1B w/o POA & 47.9  & 6.3 & 18.5 \\ 
        $A_0$-1B w/o SIAL & 61.1  & 10.2& 19.6\\  \bottomrule
    \end{tabular}
    \vspace{-0.3cm}
\end{table}


\begin{figure*}[!ht]
\begin{center}
\includegraphics[width=0.9\linewidth]{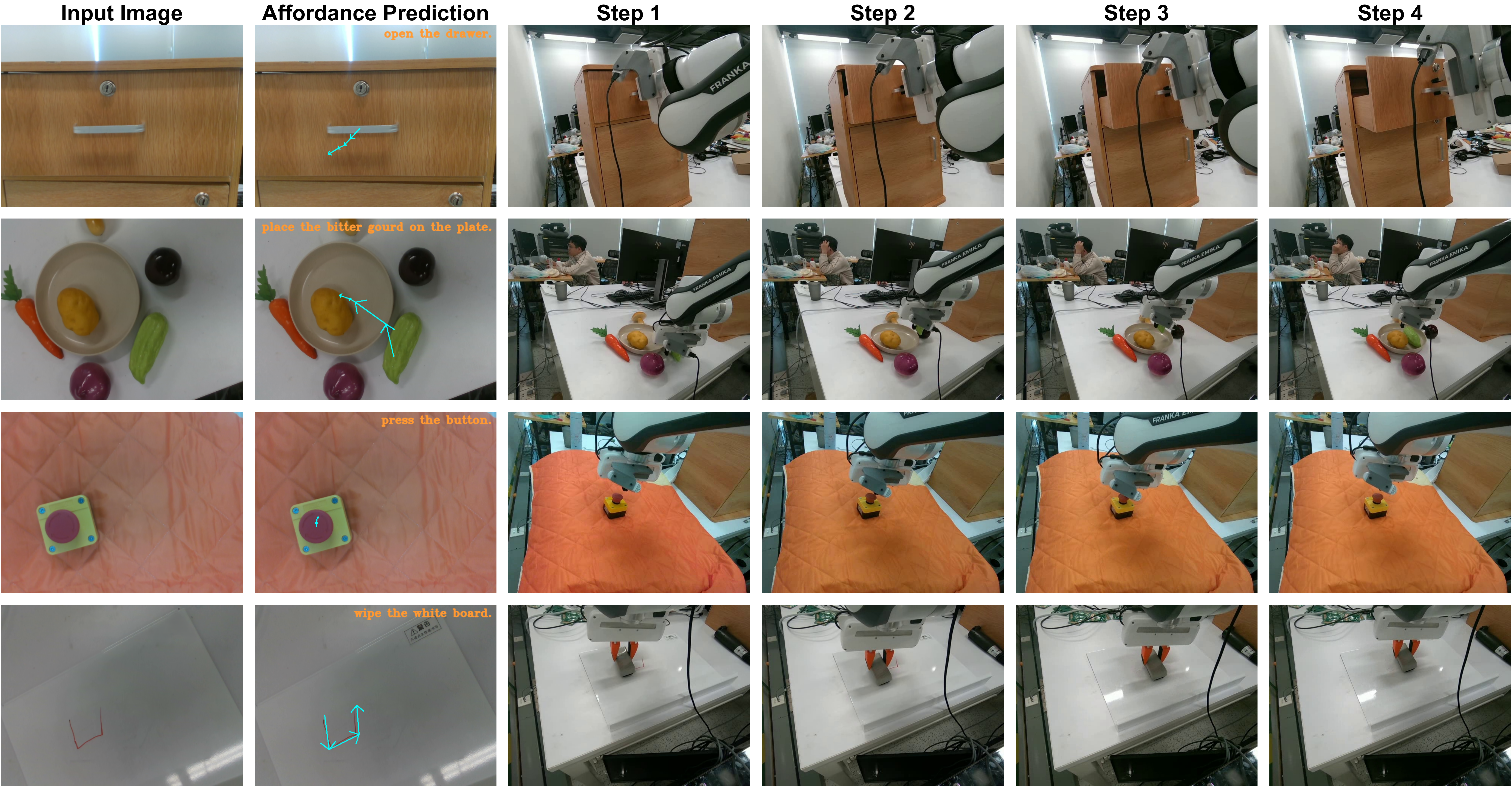}
\vspace{-3mm}
\end{center}
   \caption{Evaluation on a range of complex and temporally extended tasks using the Franka Emika robot. The four tasks include opening a drawer, placing an object on a plate, pressing a button, and wiping a whiteboard. We predict 2D affordances and employ the action execution method to deploy them on the robot.   }
\label{fig:demo_franka}
\vspace{-3mm}
\end{figure*}

\begin{figure}[!h]
\begin{center}
\includegraphics[width=0.9\linewidth]{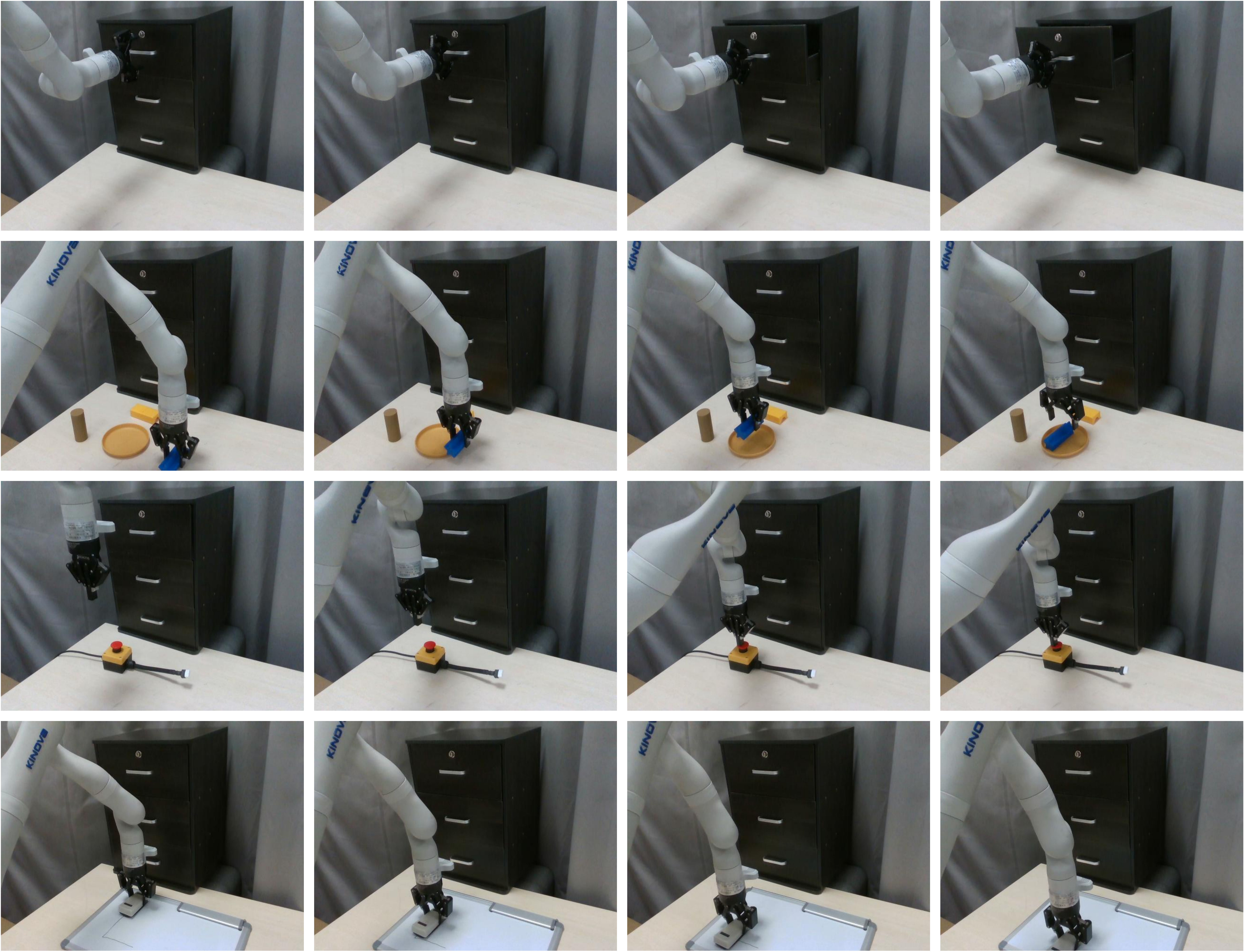}
\vspace{-3mm}
\end{center}
   \caption{Qulitative evaluation of 4 tasks on Kinova. Our model performs single-shot inference, allowing for rapid execution. }
\label{fig:demo_kinova}
\vspace{-3mm}
\end{figure}


\begin{table*}[h]
    \centering
    \caption{Performance evaluation of different large language model-based policies and Vision-language-action policies across four manipulation tasks on two distinct robotic platforms. Our method demonstrates strong platform-agnostic capabilities by achieving consistently high success rates across both Kinova Gen3 and Franka Emika robots.}
    \label{tab:manipulation_results}
    \renewcommand{\arraystretch}{1}
    \begin{tabular}{llccccc}
        \toprule
        Robot & Method & Place Object & Open Drawer & Press Button & Wipe Board & Avg. Success \\
        \midrule
        \multirow{3}{*}{Kinova } & MOKA  & 70 & 50 & 30 & 30 & 45.00 \\
                                 & ReKep  & \textbf{75} & 55 & 5  & 0  & 33.75 \\
                                 & RDT & 20 & 0 & 25 & 0 &11.25 \\
                                  \rowcolor{lightgray} & $A_0$-1B     & 60 & \textbf{65} & \textbf{40} & \textbf{50} & \textbf{53.75} \\
        \midrule
        \multirow{3}{*}{Franka } & Magma  & 25 & 10 & 30 & 0  & 16.25 \\
                                 & Molmo  & 60 & 40 & 55 & 20 & 43.75 \\
                                 \rowcolor{lightgray}&  \textbf{$A_0$}-1B & \textbf{60} & \textbf{75} & \textbf{70} & \textbf{45} & \textbf{62.50} \\
        \bottomrule
    \end{tabular}
    \vspace{-0.2cm}
\end{table*}

\subsection{RealWorld Experiments}
\label{sec:realworld}
In this section, we evaluate $A_0$'s performance in real-world environments, comparing with state-of-the-art 2D affordance-based methods, MOKA\cite{liu2024moka}, ReKep~\cite{huang2024rekep} and end-to-end approaches, $\pi_0$~\cite{black2024pi_0}, RDT~\cite{liu2024rdt}. Our experiments were conducted on multiple robotic platforms including Franka Emika, Kinova Gen3, Realman, and Dobot X-Trainer to validate generalization capabilities across different embodiments.

\subsubsection{Tasks Setting}
For all experiments, we used a standardized environment setup with consistent object placements and camera configurations across all methods. Each robot was equipped with an RGB-D camera for visual perception. As shown in Figure \ref{fig:demo_franka}, the testing protocol involved four common household tasks: (a) open the drawer, (b) placing an object on the plate, (c) press the button, (d) and wipe the white board. In the following sections, these tasks are abbreviated as Place Object, Open Drawer, Press Button, and Wipe Board. For all tasks, we conducted 20 trials to calculate the success rate.
Our approach takes an input image along with the corresponding instruction. After passing through $A_0$, it generates a 2D affordance prediction (shown in the second column of the figure). Subsequently, our action execution module transforms this prediction into an action in the SE(3) space, which is then executed by the robotic arm (shown in the right 4 columns of the figure).

\begin{table*}[!h]
    \centering
    \caption{Comparison with RDT-1B and $\pi_0$ on the Wipe Board task using the Kinova platform, highlighting our method's superiority in trajectory-following and task execution efficiency.}
    \label{tab:end2end}
    \begin{tabular}{l|c|c|c|c|c|c}
   \toprule
        ~ & Place Object & Open Drawer & Press Button & Wipe Board & Steps & Avg. Success\\ \midrule
        RDT-1B~\cite{liu2024rdt} & 20 & 0 & 25 &  0 & 11.25  & 25-50\\ 
        $\pi_0$~\cite{black2024pi_0} & 40 &20 &10 & 10 &20.00 & 25-50 \\ 
        $\pi_0$ + FAST~\cite{black2024pi_0} & 35 & 10 &30 &0 &18.75  & 25-50\\ 
        \rowcolor{lightgray}$A_0$-1B & \textbf{60} & \textbf{65} & \textbf{40} & \textbf{50} & \textbf{53.75} & \textbf{4-5} \\ \bottomrule
    \end{tabular}
    \vspace{-0.2cm}
\end{table*}

\subsubsection{Compare with 2D Affordance Methods}


We compare our approach with two categories of methods: 1) MOKA~\cite{liu2024moka} and ReKep~\cite{huang2024rekep}, which utilize vision-based models to predict 2D affordances, and 2) Molmo~\cite{deitke2024molmo} and Magma~\cite{yang2025magma}, which predict 2D points using Visual-Language Models (VLM) outputs and execute actions via our action execution module. The first category demonstrates the superior ability of our model in \textbf{spatial affordance understanding}, while the second category shows that \textbf{action execution module} can be seamlessly integrated into other methods, exhibiting strong generalization capabilities. 
We reproduced MOKA and ReKep on the Kinova platform. For Magma and Molmo, we obtained 2D points by feeding the camera input and prompts into their models, then executed the actions using our action execution module.
The results are shown in Table \ref{tab:manipulation_results}. Notably, it excels in trajectory-based tasks such as Wipe Board, where precise motion execution is critical.
Our approach achieves an average success rate of 62.50\% on Franka, outperforming the next best method (Molmo: 43.75\%) by 18.75
percentage points. On Kinova, our method achieves 53.75\%, showing a 20 percentage point improvement over the weakest baseline
(ReKep: 33.75\%). Particularly in the Open Drawer task, our model achieves 75\%, surpassing all competitors, and in the Wipe Board task,
it reaches 45\%, demonstrating significant robustness in trajectory tracking. Our method performs slightly weaker than MOKA and ReKep on the Place Object task on the Kinova. A possible reason for this is that these methods, based on vision foundation models like SAM~\cite{kirillov2023segany} and GPT-4~\cite{2023GPT4VisionSC}, have encountered a vast number of real-world physical objects during their training. We also compare with Robopoint. Details are in supporting materials.

\begin{figure}[!h]
\begin{center}
\includegraphics[width=1\linewidth]{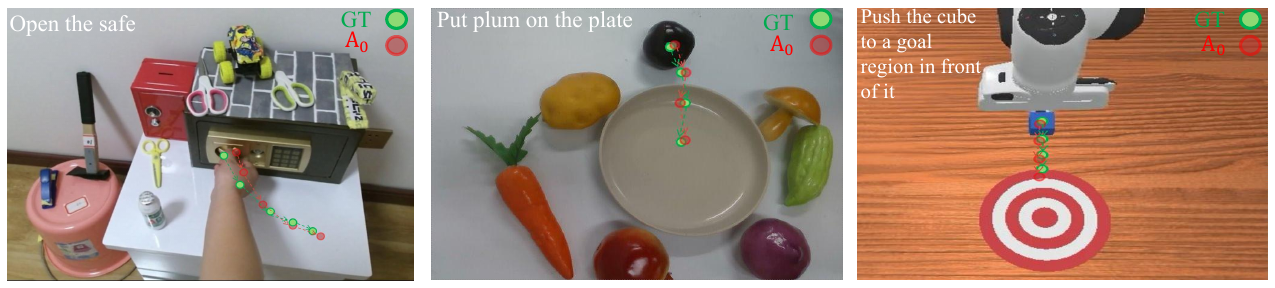}
\end{center}
   \caption{$A_0$ can predict accurately on different datasets. These three images are sourced from Droid, our Franka robot, and the ManiSkill simulation environment. }
\label{fig:predict}
\vspace{-0.6cm}
\end{figure}

\subsubsection{Compare with Vision-Language-Action Methods}
We also conducted a comparative analysis with the latest End-to-End Vision-Language-Action (VLA) large model, specifically RDT-1B~\cite{liu2024rdt} and $\pi_0$~\cite{black2024pi_0}. RDT-1B is a 1-billion-parameter Diffusion Transformer for imitation learning, pre-trained on over 1 million multi-robot episodes. RDT unifies the actions of different robots into a 128-dimensional space while supporting flexible input and output configurations.
$\pi_0$ proposes a novel flow matching architecture built on top of a pre-trained VLM to inherit Internet-scale semantic knowledge. To demonstrate the superiority of our method on trajectory-based tasks, we collected 5 episodes image-action data for each task, which includes third-view images and wrist images. We fine-tuned these two models and conducted experiments on the single-arm Kinova robot. As shown in Table~\ref{tab:end2end}, our method significantly outperforms prior VLA-based methods, achieving an average success improvement of \textbf{33.75\%} over the second-best method. Notably, on the most trajectory-intensive task ``Wipe Board", our method improves success rate by \textbf{40\%}.
In terms of execution efficiency, VLA methods like RDT-1B and $\pi_0$ rely on step-by-step model inference, typically requiring 25--50 steps to complete a task. In contrast, our method predicts 4--5 key waypoints at the beginning, resulting in significantly fewer execution steps and faster rollout. This distinction is reflected in the ``Steps'' column of Table~\ref{tab:end2end}.


\subsubsection{Qualitative Evaluation}
In Figures \ref{fig:demo_franka} and \ref{fig:demo_kinova}, we present 8 examples of our method. The method in Figure \ref{fig:demo_franka} is performed on the Franka platform, while the method in Figure \ref{fig:demo_kinova} is executed on the Kinova platform. The image sequences of task execution in these 8 examples effectively demonstrate that our method is platform-agnostic. Since our method is based on single-shot inference, the robot can be deployed and execute tasks very quickly. We also demonstrate the performance of $A_0$ across diverse datasets in Figure \ref{fig:predict}.

\section{Conclusion}

 In this paper, we present $A_0$, a hierarchical affordance-aware diffusion model for robotic manipulation. By decomposing tasks into high-level spatial affordance reasoning and low-level action execution, $A_0$ uses an Embodiment-Agnostic Affordance Representation to predict object-centric contact points and trajectories, enabling generalization across robotic platforms. Pre-trained on 1 million contact points and fine-tuned on annotated trajectories, $A_0$ outperforms state-of-the-art methods in tasks like wiping and stacking. Key components like Position Offset Attention and Spatial Information Aggregation enhance spatial reasoning and efficiency. Experiments on Franka, Kinova, Realman, and Dobot validate $A_0$'s robustness and real-world applicability. For long-horizon planning and orientation-sensitive tasks, VLMs are leveraged to decompose the task, which can then be executed by $A_0$ stage by stage. Future work will extend $A_0$ to dynamic and unstructured environments.

\appendix
\phantomsection
\maketitlesupplementary

\section{Social Impact}
The internet datasets we used, PixMo-Points, HOI4D, and the real-robot dataset Droid, are all publicly available and transparent. The ManiSkill-5k dataset was collected in a simulation environment, while the real-robot datasets we collected do not contain any personal information. We plan to release these datasets in the future. Our method has no ethical risk on dataset usage and privacy violation since all the benchmarks are publicly available and transparent.

\section{Limitations and Future work}

\subsection{Limitaions}
Our method has two main limitations:
\begin{itemize}
\item Our model relies on methods like gripper samplers to predict grasp poses for action execution. However, existing approaches in this area often exhibit suboptimal performance and limited generalization across different tasks.
\item Our method requires a depth map to estimate height, followed by refinement using a VLM. However, this approach may not perform well in tasks involving occluded objects.
\end{itemize}

\subsection{Discussion}
\begin{itemize}
\item \textbf{Orientation-sensitive tasks.}
Our method focuses on high-level spatial understanding. The action execution module is fully modular and can be replaced.
For tasks requiring precise orientation and nuanced 6D manipulation—such as liquid pouring and revolute‐drawer opening—We will choose an optimal observation viewpoint that allows $A_0$ to predict the necessary waypoints. As demonstrated in prior work (e.g., ATM: Any-point trajectory modeling \cite{WenLS0D0A24}), a track‐guided policy can be trained with only a few demonstrations to achieve accurate control. For more complex tasks, VLMs like GPT-4o can be employed to decompose the task into a sequence of subtasks and $A_0$ can address each step individually. 
\item {\textbf{Long-horizon planning.}} 
Our method faces this common limitation shared by affordance-based and modular approaches. Even current VLA models struggle with long-horizon tasks. In future work, we will address this by leveraging VLMs such as GPT-4o to decompose long-horizon tasks into a sequence of shorter subtasks (e.g., $\pi_{0.5}$), which can then be executed stage-by-stage using $A_0$.
While MOKA and ReKep handle multiple objects via prompt VLM, we have incorporated VLM for high-level planning in our new experiments. This enables our model to perform more complex tasks, such as inserting flowers into the bottle and putting  fruits on the plate, as shown in Fig. \ref{fig:demo_kinova}.

\end{itemize}

\subsection{Future Work}
Based on our analysis of the limitations of our method, we plan to pursue the following future work:
\begin{itemize}
    \item First, improving grasp pose estimation: Our current method relies on a gripper sampler to obtain grasp poses. A potential improvement could be leveraging a VLM to visually assist in selecting the best gripper position from gripper candidates or directly prompting the VLM to generate a grasp pose.
    \item Second, improving height estimation: Currently, the grasp height is obtained by prompting an LLM. We can refine our model by incorporating depth, gripper length, and other relevant information as conditions to directly predict the height.
\end{itemize}

\begin{figure}[t]
\begin{center}
\includegraphics[width=1.0\linewidth]{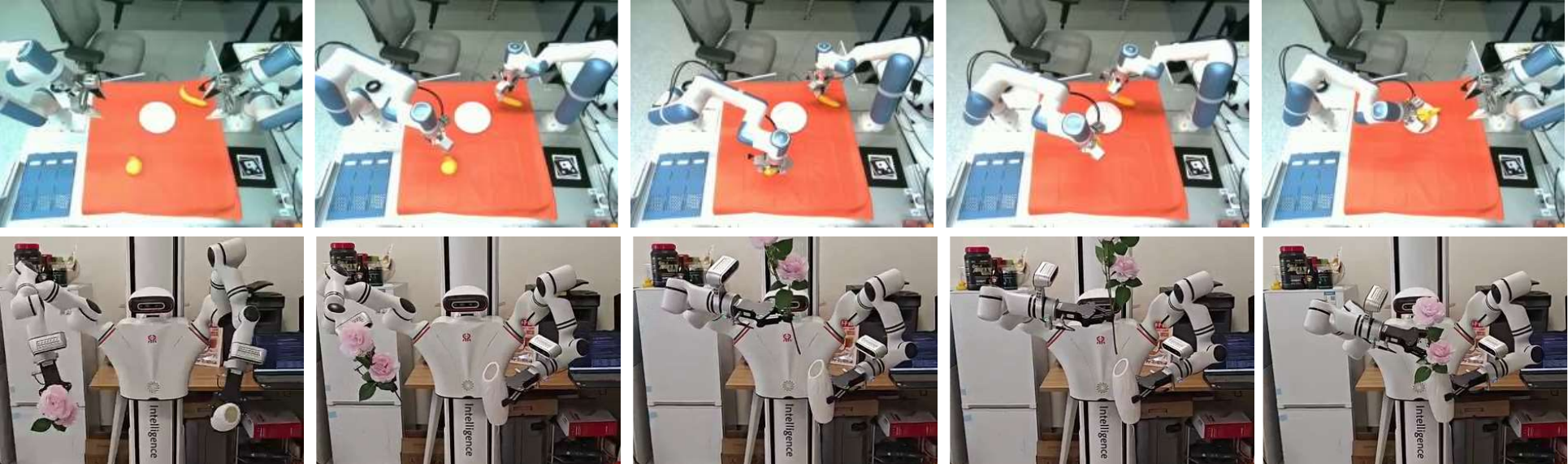}
\end{center}
   \caption{We incorporate a VLM for high-level planning, which enables our model to perform more complex tasks. First row: put fruits on a plate. Second row: insert flowers into the bottle. }
\label{fig:demo_kinova}

\end{figure}

\section{Dataset}
We have collected data from four types of sources: internet data, real-robot data, human-centric data, and simulator data. Below is a detailed description of each of these four types of data.

\textbf{Contact-point Localization Data:}
We select one million samples with only a single coordinate point (contact point \(c_0^{2\text{D}}\)) annotation from the PixMo-Points \cite{deitke2024molmo} dataset, named PixMo-One-Point. Every sample consists of one image, object label and corresponding coordinate.

\textbf{Real Robotic Data:}
We developed a multi-stage annotation pipeline to capture high-quality manipulation trajectories from real robotic interactions. Human annotators identified target objects and initial contact points in videos, followed by automatic tracking using the CoTracker model \cite{karaev2023cotracker} to generate trajectory waypoints. In the second method, we employed Molmo \cite{deitke2024molmo} to annotate the initial point and subsequently used SAM2 \cite{ravi2024sam} to segment and track the object mask across frames. Each trajectory was manually verified for accuracy. This semi-automated approach produced a dataset of verified manipulation waypoints. We randomly selected and annotated 3,056 trajectory samples from the DROID \cite{khazatsky2024droid} dataset, naming it DROID-3k.

\textbf{Human-centric Data:}
Compared to robot interaction data, video-based human-object interaction data like HOI4D~\cite{liu2022hoi4d} are more accessible and semantically rich. 
The HOI4D dataset \cite{liu2022hoi4d} encompasses 16 distinct object categories (e.g., toy cars, bottles) and covers 6 different tasks, including pick-and-place, opening a drawer, and pulling a toy car. Each category includes multiple videos, amounting to a total of 3,572 videos. Each video is comprised of several segmented actions, and by employing various action labels to delineate these segments, a total of 22,140 unique video segments were obtained. We convert the original dataset into a 2D waypoint format, where the center point of the object’s 2D mask in each frame is used as the waypoint, and the combination of the action and object name serves as the instruction. We refer to the transformed dataset as HOI4D-22k.

\textbf{Simulator Data:}
To adapt our model for various deployment environments, we collected 4965 trajectories from the ManiSkill Scene dataset, converting 3D data to 2D for compatibility. There are five tasks: “Peg Insertion Side”, “Plug Charger”, “Pull Cube Tool”, “Push Cube”, and “Stack Cube”. Each task contains about a thousand trajectories. We named this dataset Maniskill-5k. 
Camera angles were adjusted to diversify scene data.

The image resolutions of the HOI4D, Maniskill, and DROID datasets are 1920×1080, 512×512, and 320×180, respectively.



\begin{figure*}[t]
    \centering   \includegraphics[width=1.0\linewidth]{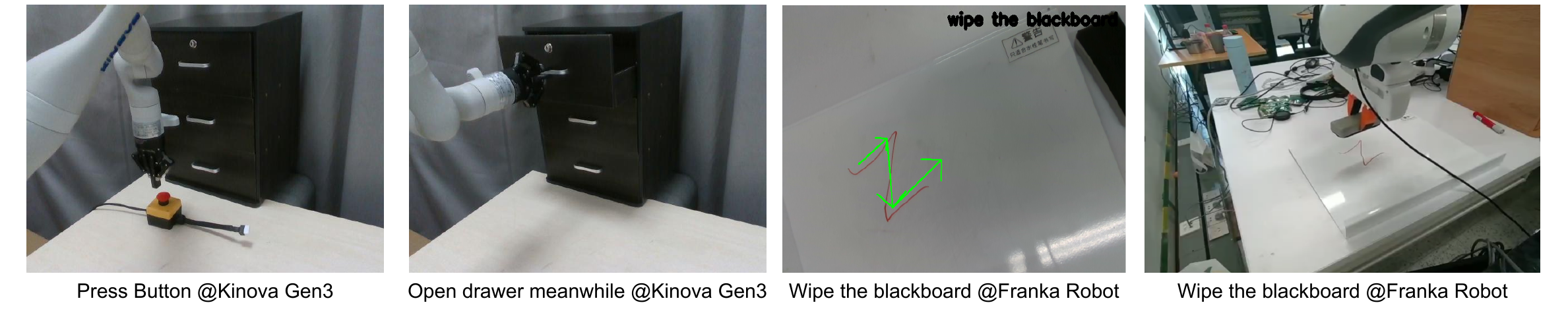}
    \caption{Real World Kinova Gen3 Robot}
    \label{fig:appendix-1}
\end{figure*}

\begin{figure*}[t]
    \centering   \includegraphics[width=1.0\linewidth]{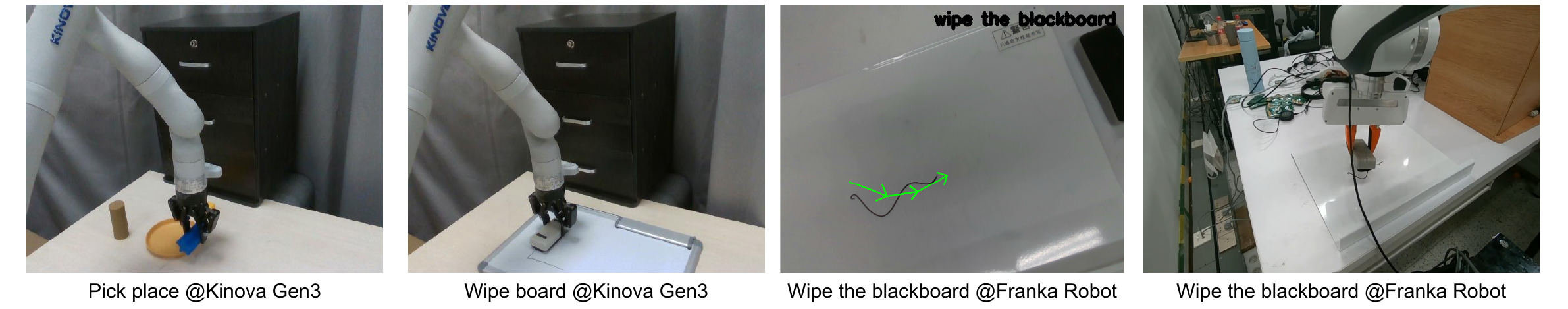}
    \caption{Real World Franka Robot}
    \label{fig:appendix-2}
\end{figure*}

\begin{figure}[t]
    \centering   \includegraphics[width=1\linewidth]{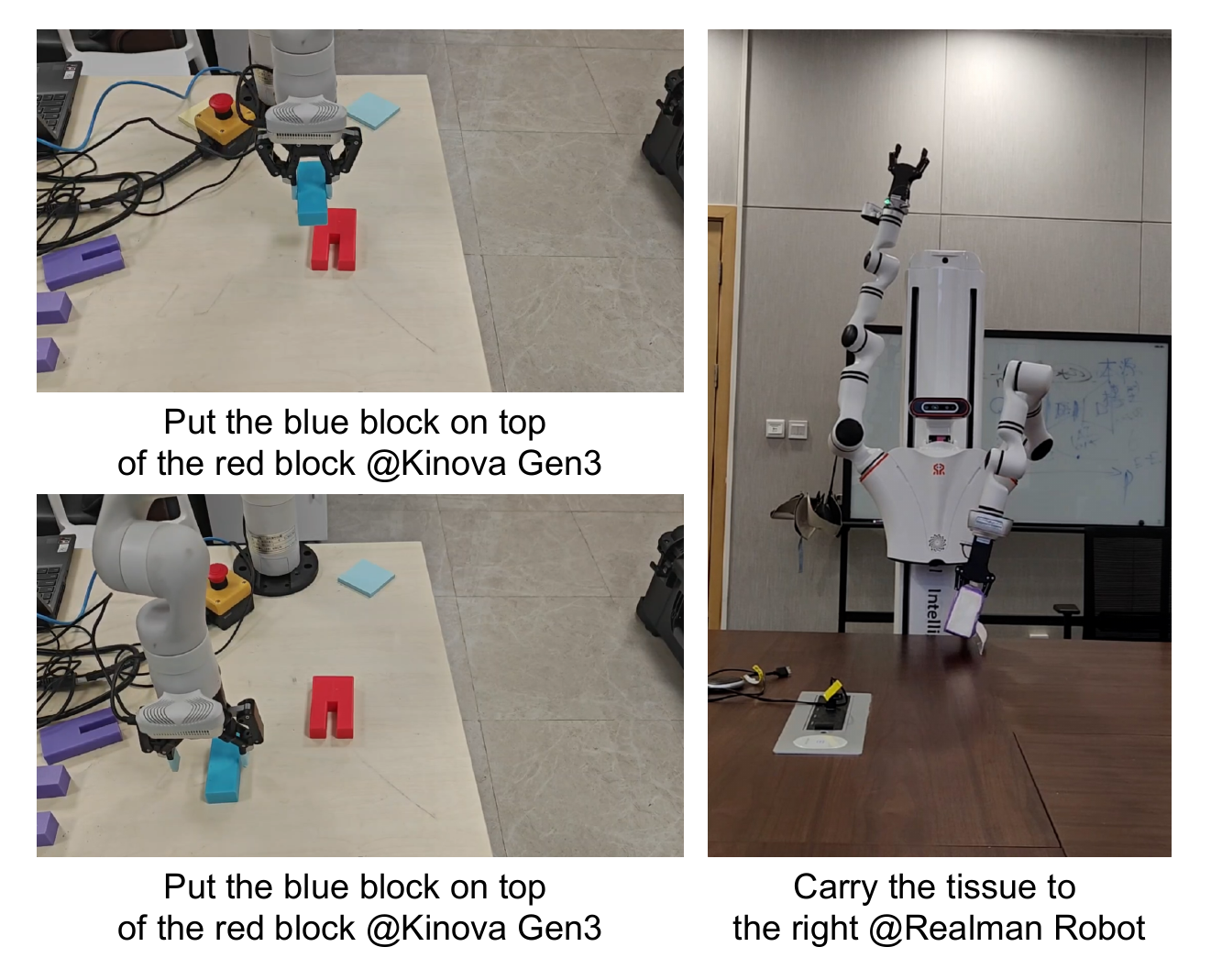}
    \caption{Real World Kinova Gen3 Robot and Realman Robot}
    \label{fig:appendix-3}
\end{figure}

\section{Compare with Robopoint}

To compare our approach with Robopoint, we evaluated both methods on two standard datasets: HOI4D and DROID. We measured the Mean Absolute Error (MAE) of the first predicted interaction pixel compared to the ground truth. For our model, we used the first waypoint directly, while for Robopoint, which predicts interaction regions rather than specific points, we used the average position of all predicted points for comparison.

As shown in Table 3, our method achieves substantially lower MAE scores across both datasets. Specifically, on HOI4D, our model achieves an MAE of 54.46 compared to Robopoint's 121.09, representing a 55.2\% reduction in error. Similarly, on DROID, our approach attains an MAE of 14.13 versus Robopoint's 27.47, a 40.4\% improvement. These results demonstrate that our method provides more precise interaction point predictions, which is crucial for accurate robot manipulation tasks.

\begin{table}[!ht]
    \centering
    \begin{tabular}{l|cc}
    \toprule
    Method & HOI4D (MAE) & DROID (MAE) \\ \midrule
    Robopoint & 121.09 & 27.47 \\ 
    Ours & \textbf{54.46} & \textbf{14.13} \\ \bottomrule
    \end{tabular}
    \caption{Comparison of Mean Absolute Error (MAE) for the first interaction pixel between our method and Robopoint on HOI4D and DROID datasets. Lower values indicate better performance.}
    \label{tab:robopoint_comparison}
\end{table}

\renewcommand{\thesection}{\Alph{section}} 


\section{Real World Experiment}

We conduct our real world robot experiments on multiple robot platforms, including Kinova, Franka, and Realman Robot, as shown in \Cref{fig:appendix-1}, \Cref{fig:appendix-2} and \Cref{fig:appendix-3}.

\section{Annotation platform}

Our annotation platform is a multi-source, semi-automated system designed to generate high-quality spatial affordance data for robotic manipulation. It integrates real robotic interactions, human-object interaction datasets, simulation environments, and large-scale internet-sourced datasets to create a standardized Embodiment-Agnostic Affordance Representation.

\begin{figure*}[t]
    \centering   \includegraphics[width=1.0\linewidth]{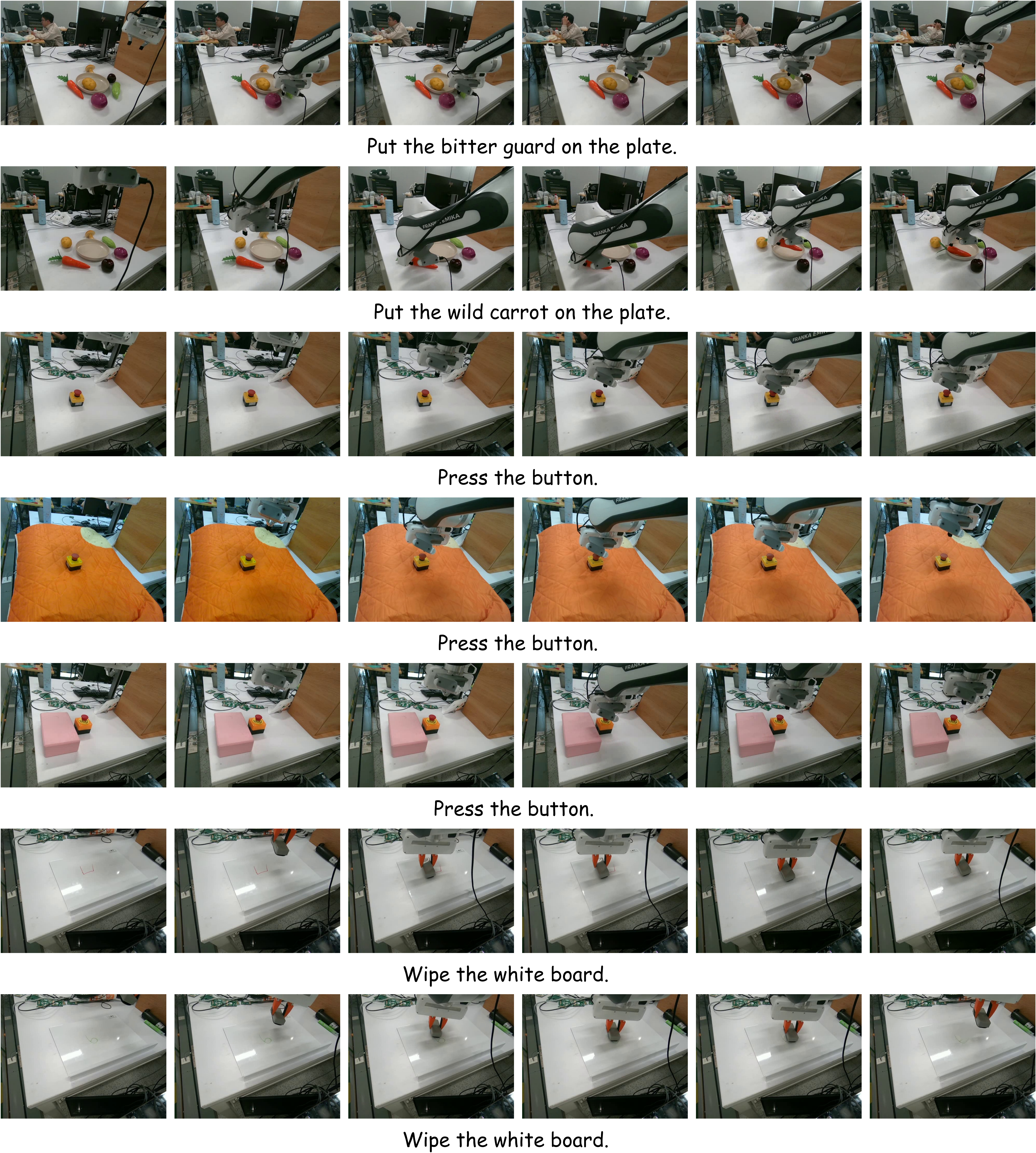}
    \caption{Real World Franka Robot.}
    \label{fig:appendix_franka_add}
\end{figure*}
\section{Additional Qualitative Result on Franka}
As shown in Fig. \ref{fig:appendix_franka_add}, our approach generalizes well to different backgrounds on Franka Robot. The text below each raw of images serves as the instruction.

{
    \small
    \bibliographystyle{ieeenat_fullname}
    \bibliography{main}
}

\end{document}